\documentclass[french,12pt,titlepage]{article}
\usepackage{geometry}
\usepackage[utf8]{inputenc}
\usepackage[T1]{fontenc}
\usepackage{bbm}
\usepackage{babel}
\usepackage{amssymb}
\usepackage{appendix}
\usepackage{subfigure}
\usepackage{graphicx}
\graphicspath{ {images/} }
\usepackage{booktabs}

\title{Optimisation de la recette dans le cas d'une demande dépendante et sensible au prix}
\author{
        Julien Laasri\thanks{\'Ecole polytechnique, Route de Saclay, 91128 Palaiseau, France}
            \and
        Marc Revol\footnotemark[1]
}
\date{2018}
\begin{document}
\maketitle

\newpage
\section*{Remerciements}
Nous remercions Alexandre Pizzut et Pierre de Fréminville de la compagnie Air France pour nous avoir fourni l'ensemble des données rendant cette étude possible ainsi que pour nous avoir aidé régulièrement tout au long de l'avancement de notre projet.
\newpage
\tableofcontents
\newpage
\section{Introduction}

Comme décrit par Kalyan T. Talluri et Garrett J. Van Ryzin dans leur ouvrage \cite{livreRM}, le Revenue Management consiste en la maximisation du revenu d'un organisme à partir de trois types de catégories de décision :
\begin{itemize}
    \item \textbf{Les décisions structurelles} : La manière de vendre un produit (prix donnés, enchères, négociations, \ldots), les conditions d'achat (possibilité de rembourser, réduction au delà d'un certain volume, \ldots), la manière de regrouper des produits, etc.
    \item \textbf{Les décisions de prix} : Comment fixer les prix, comment prendre en compte les caractéristiques individuelles des potentiels clients, comment faire varier les prix à travers le temps, etc.
    \item \textbf{Les décisions de quantité} : Quand accepter ou rejeter une offre d'achat, Protection d'un produit pour une vente future, etc.
\end{itemize}

\begin{figure}[!h]
\centering
\includegraphics[width=15.0cm]{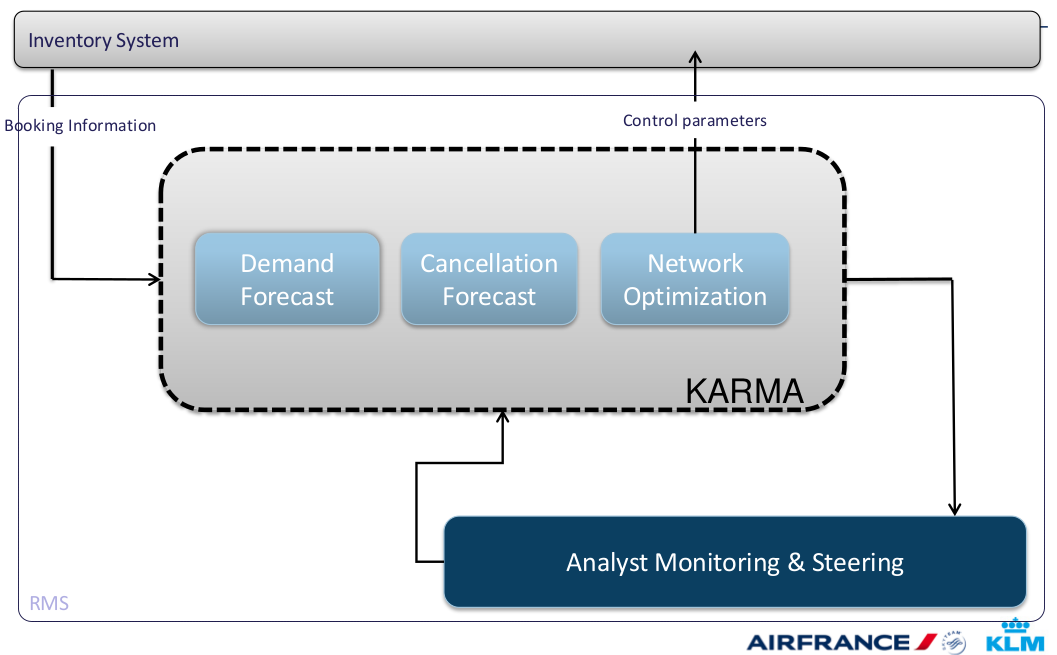}
\caption{Le Revenue Management chez Air France}
\label{graph-rev-man}
\end{figure}

Dans ce document, nous nous intéresserons principalement aux décisions de type prix et quantité pour la vente de billets d'avion sur un vol direct au cours d'un certain nombre de pas de temps. Plus précisément, nous nous situerons dans la partie optimisation de la Figure \ref{graph-rev-man}. Nous prendrons ainsi les données de demande comme acquises, car elles sont estimées au préalable par Air France à partir des données réelles. De même, pour chaque produit que l'on cherchera à vendre, on nous impose en amont les prix possibles que l'on a droit d'utiliser et qui se basent sur des algorithmes d'Air France dont les résultats sont vérifiés par des analystes. Notre but sera alors de maximiser le revenu d'un vol direct en choisissant les prix de chaque produit parmi ceux imposés.

\section{Etat de l'art et modèles classiques}\label{modeles classiques}
Dans cette section nous présentons les modèles les plus connus en Revenue Management. Nous allons nous servir de ces modèles comme base pour élaborer des modèles plus complexes. 

\subsection{Hypothèses principales}\label{hypotheses}
La plupart des modèles classiques s'appuient sur un même jeu d'hypothèses permettant de réduire la complexité du problème à résoudre. Ils offrent tous une solution qualifiée de statique au problème du leg (i.e. du vol direct) visant à maximiser le revenu moyen.

\begin{enumerate}
\item La demande pour chaque classe est indépendante 
\item La demande pour chaque classe arrive dans des intervalles de temps disjoints ordonnés par tarif croissant ( Les premiers arrivés sont intéressés par la classe la moins chère)
\item La demande pour chaque classe est modélisée par des variables aléatoires indépendantes
\item Il n'y a pas de réservation de groupe ni de possibilité de sur-classement
\end{enumerate}

Ces hypothèses sont certes simplificatrices mais peuvent se justifier. Par exemple, la deuxième hypothèse constitue une approximation raisonnable car l'expérience montre que les clients à la recherche de tarifs avantageux ont tendance à s'y prendre plus tôt que ceux des classes affaires qui réservent au dernier moment. De plus, on peut considérer qu'il s'agit d'une stratégie envisageant le pire scénario possible. En effet, si une personne d'une classe plus chère arrive plus tôt, on lui vend dans tous les cas sa place puisqu'elle rapportera davantage.
La seule hypothèse difficilement justifiable est la première. En effet, à moins que les classes proposées ne soient réellement distinctes, il existe des dépendances entre les classes. Si un client recherche une certaine classe et en trouve une autre équivalente mais moins cher, il peut être tenter d'économiser de l'argent en modifiant son achat initialement prévu. A l'inverse, s'il n'y a plus de place dans la classe qu'il recherchait, il peut être contraint d'acheter dans une autre classe à un tarif plus élevé. Ces effets, respectivement nommé buy-down et buy-up, ne sont pas pris en compte dans la modélisation classique. Nous verrons dans la partie suivante comment les incorporer dans nos modèles.

    \begin{figure}[!h]
        \begin{center}
          \includegraphics[width=12.0cm]{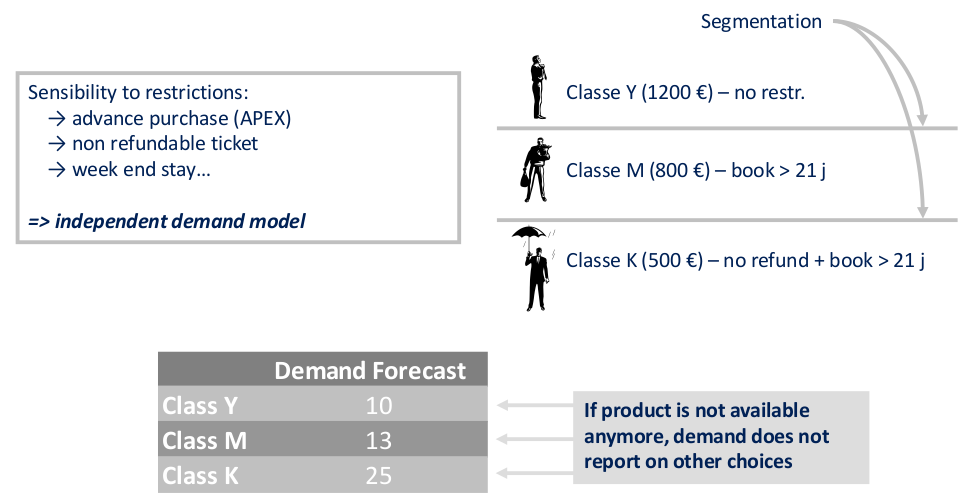}
          \label{fig:su3}
        \end{center}
        \caption{Illustration de la segmentation entre les classes}
        \label{fig:slide23}
    \end{figure}

\subsection{Le modèle à deux classes de Littlewood}\label{hypotheses}
Le modèle le plus simple en Revenue Management est dû à Littlewood \cite{livreRM}. On considère dans ce modèle deux classes $C_1$ et $C_2$ ayant chacun un prix $p_1$ et  et une demande estimée $D_1$ et $D_2$. On suppose que les hypothèses classiques évoquées dans le paragraphe précédent sont vérifiées. On a $p_1 > p_2$ et la demande $D_1$ arrive après $D_2$. La demande $D_j$ est donc stochastique et caractérisée par sa fonction de répartition $F_j(.)$. L'objectif est de maximiser le revenu moyen. Compte tenu des hypothèses, cela revient à déterminer la quantité $y^*$ de passagers de la classe $C_2$ à accepter avant de la fermer pour sécuriser le reste des sièges pour des passagers de la classe $C_1$ rapportant davantage.

\paragraph{}\label{resolutionlittle}
Le problème peut se résoudre par une simple étude des revenus marginaux. Supposons qu'il nous reste $x$ places disponibles et qu'une demande pour $C_2$ se manifeste. En acceptant la transaction, on collecte un revenu $p_2$. En la refusant, on gagne $p_1$ si et seulement si il y a assez de demande pour $C_1$ soit $D_1 \geq x$. La transaction est donc avantageuse si $p_2 > p_1 \times \mathbbm{1}_{D_1 \geq x}$, ce qui nous donne en passant à l'espérance la condition d'acceptation: $p_2 > p_1  \mathbb{P}(D_1 \geq x)$
Le terme de droite étant décroissant, on va pouvoir trouver une protection optimale $y^* = F_1^{-1}(1 - \frac{p_2}{p_1})$. La demande est ici supposée continue mais on peut appliquer la règle de Littlewood en prenant la partie entière du résultat.

\subsection{Généralisation à N classes}\label{nclasses}
Dans le cas plus général de $n$ classes, on considère toujours qu'il existe des tarifs $p_1 > p_2 > \ldots > p_n$ tels que toute la demande de la classe $n$ va arriver avant celle de la classe $n-1$ et ainsi de suite. La résolution du problème est alors plus complexe et il existe principalement deux méthodes: la programmation dynamique et la mise en place d'une heuristique se basant sur le modèle à deux classes. La première ne nous intéressera pas dans le cadre de ce projet car elle ne pourrait pas se généraliser au cas plus complexe du réseau (plusieurs vols à optimiser). Elle s'appuie sur la récurrence intuitive suivante: En notant $V_j(x)$ le revenu en ne considérant que les $j$ classes les moins chères pour une capacité restante $x$, on a:  
$$V_j(x) =  \mathbb{E}\left(\displaystyle \max_{0\leq u \leq \min{(D_j,x)}} (p_ju + V_{j-1}(x-u))\right)$$

\paragraph{L'heuristique EMSRb}\label{emsrb}
L'heuristique la plus populaire est l'EMSRb (expected marginal seat revenue version b). L'idée consiste à agréger les demandes futures afin de se ramener au problème à deux classes de Littlewood. Plus précisément, la demande future est traitée comme une unique classe ayant un revenu égale à une moyenne pondérée des revenus des classes la composant. Considérons l'agrégation des demandes futures $S_j = \sum \limits_{k=1}^j D_k $, le temps étant décroissant, ainsi que le revenu moyen des classes $1,...,j$ :
$$\overline{p}_j = \frac{\sum \limits_{k=1}^j p_k\mathbb{E}(D_k)}{\sum \limits_{k=1}^j \mathbb{E}(D_k)} $$
Le nombre de sièges $y_j^*$ à protéger pour la classe $j$ est alors donné par :
$$\mathbb{P}_j(S_j \geq y_j^*) = \frac{p_{j+1}}{\overline{p}_j}$$

\paragraph{Remarque}
Dans la pratique, on modélise souvent les $\mathbb{P}_j$ par des gaussiennes.

\subsection{Limite des modèles classiques}\label{limites}
\begin{figure}[!h]
        \begin{center}
          \includegraphics[width=15.0cm]{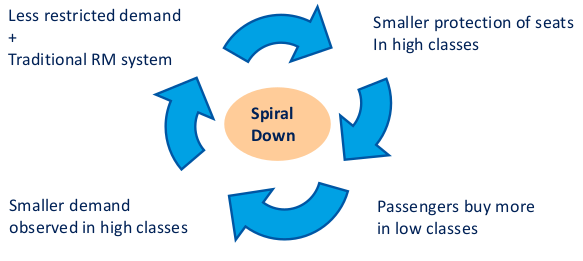}
          \label{fig:s34}
        \end{center}
        \caption{Illustration du spiral down effect}
        \label{fig:slide34}
    \end{figure}

La principale limitation des modèles classique provient de l'hypothèse d'indépendance des classes qui, comme nous l'avons mentionné, n'est pas réaliste et peut conduire à des pertes importantes. L'un de ces effets néfaste se nomme le "spiral down effect". 
Cet effet se produit lorsque des hypothèses erronées sur le comportement des clients entraînent une baisse systématique des ventes de billets à tarif élevé, des niveaux de protection et des revenus au fil du temps. Si une compagnie aérienne décide du nombre de places à vendre à un tarif élevé en fonction des ventes passées à tarif élevé, et ce sans tenir compte du fait que la disponibilité des billets à tarif réduit diminuera les ventes à tarif élevé, les ventes à tarif élevé vont diminuer (effet buy-down) , ce qui se traduit par une baisse des estimations futures de la demande de tarifs élevés. Cela donne par la suite des niveaux de protection plus bas pour les billets à tarif élevé, une plus grande disponibilité de billets à prix réduit et des ventes encore plus basses des billets à tarif élevé. Le motif continue, ce qui entraîne une spirale vers le bas (figure \ref{fig:slide34}). 

\begin{figure}[!h]
    \centering

    \subfigure{%
        \includegraphics[width=8cm]{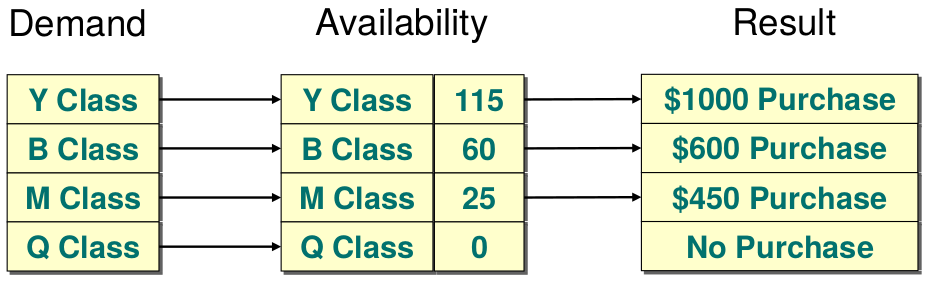}
        \label{fig:sub1}
    }
    \hfill
    \subfigure{%
        \includegraphics[width=8cm]{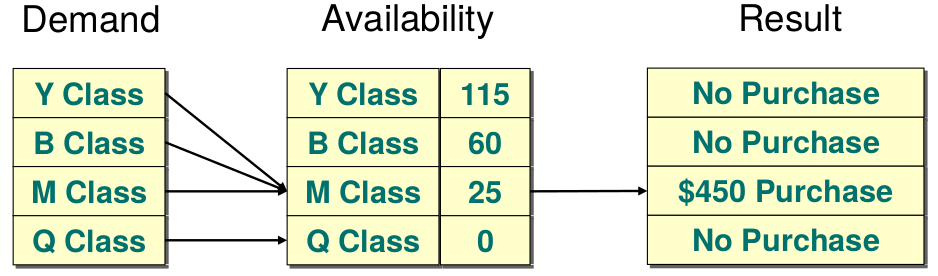}
        \label{fig:sub2}
    }
    \hfill
    \subfigure{%
        \includegraphics[width=8cm]{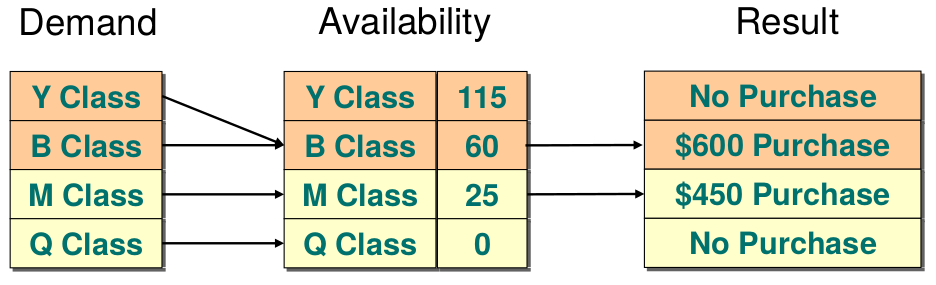}
        \label{fig:sub3}
    }

    \caption{Illustration de trois modélisations de demande. Demande segmentée (en haut à gauche), demande complètement non segmentée (en haut à droite) et le cadre hybride du regroupement par famille (en bas).}
    \label{fig:segm}
\end{figure}

\paragraph{}
L'objectif de ce projet est de proposer des modèles prenant en compte cette non-segmentation des différentes classes afin d'augmenter les revenus et d'éviter l'effet de spirale. Pour cela, nous allons dans la suite nous affranchir des hypothèses de segmentation des classes ainsi que d'ordre d'arrivée par tarifs décroissant.
Le cas de non-segmentation le plus total consiste à considérer que chaque passager peut potentiellement acheter un produit de n'importe quelle classe pourvu que celle-ci ne dépasse pas le budget qu'il s'était fixé. Si plusieurs propositions s'offrent au client, il choisira toujours la moins chère (En haut à gauche de la figure \ref{fig:segm}).

\begin{figure}[!h]
        \begin{center}
          \includegraphics[width=10.0cm]{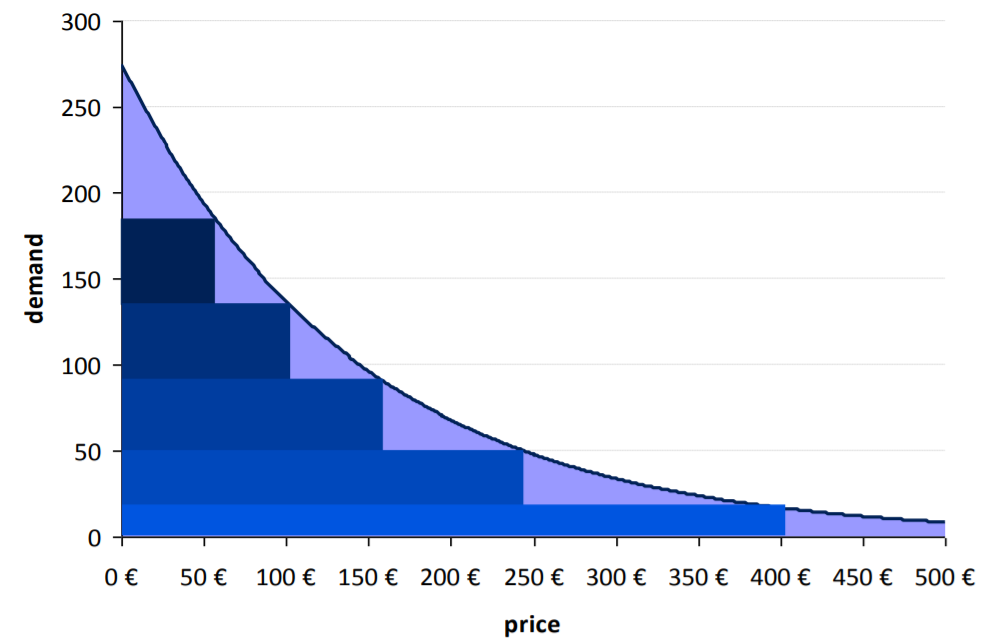}
          \label{fig:sub3}
        \end{center}
        \caption{Exemple de modélisation de l'élasticité prix de la demande}
        \label{fig:expo}
    \end{figure}

Ainsi, contrairement au cas classique, on peut modéliser la demande comme une fonction continue décroissante du prix proposé. Une modélisation consiste à considérer la relation $Q(p) = e^{-\lambda (p-p_{min})}Q_{min}$ où $p_{min}$ représente le tarif le plus bas, $Q_{min}$ la demande associée et $\lambda$ l'élasticité prix de la demande (plus lambda est important plus la demande est sensible au prix)(Figure \ref{fig:expo}). Afin de traiter un cas plus large, nous nous placerons par la suite dans le cadre d'un modèle de prédiction dit "hybride". Il s'agit d'une segmentation partielle des classes en familles (les produits) avec une segmentation entre les différentes familles (un client veut un produit donné et pas un autre) mais une absence de segmentation au sein d'une même famille (ce produit est vendu sous différentes formes à différents prix, le client achète le prix le plus bas qu'on lui propose pour son produit s'il est dans son budget) comme en bas dans la figure \ref{fig:segm}.

\section{Premier algorithme}\label{nos algorithmes}
A la suite de la réception des données sur lesquelles nous allions travailler, nous avons pu commencer à développer nos propres algorithmes.

\subsection{Modélisation mathématique du problème}\label{modelisation_mathematique}
\begin{figure}[!h]
        \begin{center}
        \subfigure{%
          \includegraphics[width=8.0cm]{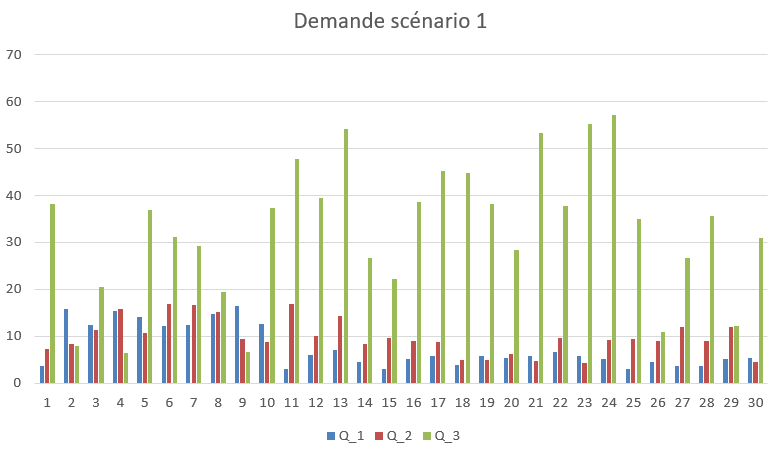}
          \label{fig:sc1}
        }
        \subfigure{%
          \includegraphics[width=8.0cm]{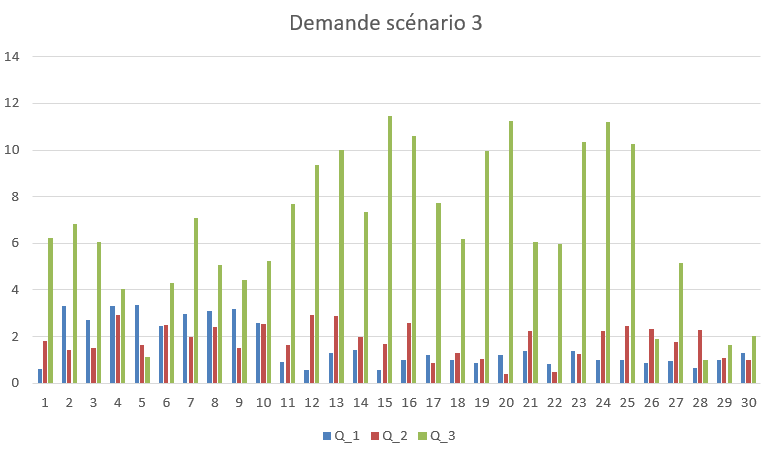}
          \label{fig:sc3}
        }
        \end{center}
        \caption{Visualisation de l'estimation de la demande dans les cas d'une demande importante (scénario 1) et faible (scénario 3)}
        \label{fig:plotsc}
    \end{figure}

Les données que nous avons reçues étaient pour un vol donné d'un point $A$ à un point $B$ sans correspondance. On avait alors accès aux données suivantes :

\begin{itemize}
\item Cinq scénarios correspondant à cinq vols différents en terme de capacité, demande et élasticité prix de la demande. (Figure \ref{fig:plotsc})
\item Trois types de produits $i$ pouvant représenter les classes Business, Economy, etc.
\item $30$ pas de temps où $t = 29$ correspond à la mise en vente des billets d'avion et $t = 0$ correspond au départ de l'avion.
\item Un ensemble de prix possibles par produit $i$ noté $\mathcal{P}_i$. On note de plus $p_i^{min} = \min_{p \in \mathcal{P}_i} p$.
\item Pour un produit $i$ donné et un temps $t$ donné, la demande moyenne $Q_{i,t}$ associée au prix le plus faible $p_i^{min}$.
\item Le FRAT5 par produit et par temps noté $F_{i,t}$. Il est défini de telle sorte que la probabilité $p_{p_i^{min}\rightarrow p}$ qu'un individu parmi les $Q_{i,t}$ initiaux achète tout de même son billet d'avion si le prix est finalement $p$ soit :
$$p_{p_i^{min}\rightarrow p} = \exp\left(-\frac{\ln(2)}{F_{i,t}-1}\,\left(\frac{p}{p_i^{min}}-1\right)\right)$$
Ainsi la demande en moyenne au temps $t$ pour le produit $i$ si on le propose au prix $p_{i,t} \in \mathcal{P}_i$ est donnée par :
$$q_{i,t}=Q_{i,t}\,\exp\left(-\frac{\ln(2)}{F_{i,t}-1}\,\left(\frac{p_{i,t}}{p_i^{min}}-1\right)\right) = \alpha_{i,t}\,\exp(-\beta_{i,t}\,p_{i,t})$$
avec $\alpha_{i,t}, \beta_{i,t} > 0$.
On peut aussi interpréter le FRAT5 par le fait que si le prix minimum $p_i^{min}$ est multiplié par $F_{i,t}$ alors la demande moyenne $q_{i,t}$ représente la moitié de la demande moyenne initiale $Q_{i,t}$.
\item La capacité de l'avion notée $C$.
\end{itemize}

On suppose dans cette partie que les résultats en moyenne sont les vraies données réelles, et que l'on peut ainsi connaître la demande future lorsque l'on fixe le prix des produits. Ainsi, en supposant tous les prix $p_{i,t}$ fixés, on a accès à la demande totale associée à ces prix, qui est la somme des $q_{i,t}$ ainsi qu'au revenu associé à cette demande, à savoir la somme des $p_{i,t}\,q_{i,t}$.

La maximisation du revenu moyen pour ce vol en tenant en compte la contrainte de capacité devient alors le problème suivant :
$$
    \max_{p_{i,t} \in \mathcal{P}_i} \sum_{i,t}p_{i,t}\,\alpha_{i,t}\,e^{-\beta_{i,t}\,p_{i,t}} \quad s.c. \,\, \sum_{i,t}\alpha_{i,t}\,e^{-\beta_{i,t}\,p_{i,t}} \leq C
$$

\subsection{Branch and Bound}\label{branch and bound}
La première idée qui nous est venue a été décrite dans le cas général par S. Gaubert et J.F. Bonnans dans leur polycopié \cite{coursRO}. Le principe est de parcourir un arbre dont chaque noeud correspond à la fixation d'un prix dans l'ensemble $\mathcal{P}$ correspondant pour un temps et un produit donné. Plus on descend dans l'arbre, plus on fixe un grand nombre de prix. La résolution du problème relâché, ici sans contrainte de bornes sur les prix, permet alors de fournir une borne sur-optimale en un point donné de l'arbre. En effet, il permet de résoudre le problème continu sur l'ensemble des prix qui n'ont pas encore été fixés lors de la descente de l'arbre. Si cette borne n'est pas suffisamment élevée par rapport aux revenus que l'on sait déjà obtenir en ayant atteint précédemment une feuille, cela ne sert à rien de poursuivre la descente de l'arbre par ce noeud.
Même si la relaxation complète des prix peut paraître extrême, on verra par la suite que les prix optimaux sont très proches de ceux des $\mathcal{P}_i$, ce qui témoigne notamment de la bonne qualité de la sélection des prix faite en amont par Air France, mais aussi de la pertinence de la borne sur-optimale ainsi trouvée.

\subsubsection{Résolution du problème relâché}\label{probleme dual}

Résolvons le problème relâché suivant :
$$
    \max_{p_{i,t} \in \mathbb{R}} \sum_{i,t}p_{i,t}\,\alpha_{i,t}\,e^{-\beta_{i,t}\,p_{i,t}} \quad s.c. \,\, \sum_{i,t}\alpha_{i,t}\,e^{-\beta_{i,t}\,p_{i,t}} \leq C
$$

On va commencer par résoudre le problème dual puis montrer qu'il n'y a pas de saut de dualité pour ce problème. Le lagrangien est ici donné par :
$$
L(p,\mu) = \sum_{i,t}p_{i,t}\,\alpha_{i,t}\,e^{-\beta_{i,t}\,p_{i,t}} + \mu\,\left(C-\sum_{i,t}\alpha_{i,t}\,e^{-\beta_{i,t}\,p_{i,t}}\right)
$$
Le problème primal est donné par :
$$ \max_{p_{i,t} \in \mathbb{R}} \min_{\mu \in \mathbb{R}_+} L(p,\mu)$$
Donc le problème dual est donné par :
$$ \min_{\mu \in \mathbb{R}_+} \max_{p_{i,t} \in \mathbb{R}} L(p,\mu)$$
Un calcul simple de dérivée nous montre que $\max_{p_{i,t} \in \mathbb{R}} L(p,\mu)$ est atteint pour $p_{i,t} = \mu + \frac{1}{\beta_{i,t}}$. D'où le problème dual se ramène à trouver le minimum sur $\mathbb{R}_+$ de la fonction $f$ définie par :
$$f(\mu) = \mu\,C+\sum_{i,t}\frac{\alpha_{i,t}}{\beta_{i,t}}e^{-\beta_{i,t}\,\mu-1}$$
Cette fonction est strictement convexe car deux fois dérivable et de dérivée seconde strictement positive. Ainsi, en notant $\hat\mu$ le point d'annulation de sa dérivée, la solution du problème dual est donc $\mu^* = \hat\mu_+$. $\hat\mu$ étant le point d'annulation d'une fonction strictement croissante, on l'obtient facilement par une méthode de Newton. Par ailleurs, on remarque dans la définition de $L(p,\mu)$ que les écarts complémentaires sont donnés par $\mu\,f'(\mu)$ lorsque $p_{i,t} = \mu + \frac{1}{\beta_{i,t}}$ et sont donc nuls en $\mu^*$ car $\mu^* = \hat\mu$ ou $\mu^* = 0$. Il n'y a donc pas de saut de dualité, et les prix optimaux sont donnés par la formule $p_{i,t}^* = \mu^* + \frac{1}{\beta_{i,t}}$.

Un exemple de résultat donné par l'implémentation de cet algorithme est donné à la Figure \ref{evol_prix} pour la scénario 1. On rappelle que le temps $t$ décroît à l'approche du décollage de l'avion.

\begin{figure}[!h]
\centering
\begin{tabular}{cccc}
  \hline
  Pas de temps t & Prix produit 1 & Prix produit 2 & Prix produit 3\\
  \hline
  0 & 841.16 & 533.04 & 353.22 \\
1 & 810.25 & 511.40 & 339.31 \\
2 & 779.33 & 489.76 & 325.40 \\
3 & 748.42 & 468.12 & 311.49 \\
4 & 717.50 & 446.48 & 297.58 \\
5 & 686.59 & 424.84 & 283.66 \\
6 & 655.67 & 403.20 & 269.75 \\
7 & 624.76 & 381.56 & 255.84 \\
8 & 593.84 & 359.92 & 241.93 \\
9 & 562.93 & 338.28 & 228.02 \\
10 & 532.01 & 316.64 & 214.11 \\
11 & 501.10 & 295.00 & 200.19 \\
12 & 470.18 & 273.36 & 186.28 \\
13 & 439.27 & 251.72 & 172.37 \\
14 & 434.76 & 248.56 & 172.37 \\
15 & 430.25 & 245.41 & 172.37 \\
16 & 425.74 & 242.25 & 172.37 \\
17 & 421.23 & 239.09 & 172.37 \\
18 & 416.73 & 235.94 & 172.37 \\
19 & 412.22 & 232.78 & 172.37 \\
20 & 407.71 & 229.63 & 172.37 \\
21 & 403.20 & 226.47 & 172.37 \\
22 & 398.69 & 223.32 & 172.37 \\
23 & 394.18 & 220.16 & 172.37 \\
24 & 389.68 & 217.00 & 172.37 \\
25 & 385.17 & 213.85 & 172.37 \\
26 & 380.66 & 210.69 & 172.37 \\
27 & 376.15 & 207.54 & 172.37 \\
28 & 371.64 & 204.38 & 172.37 \\
29 & 367.13 & 201.22 & 172.37 \\
  \hline
\end{tabular}
\caption{Évolution des prix optimaux non contraints au cours du temps pour la scénario 1 avec un avion de capacité $C=180$.}
\label{evol_prix}
\end{figure}

Le revenu obtenu pour ces prix vaut $76900.39$. C'est une bonne borne maximale pour notre problème contraint car ces prix sont dans les mêmes gammes que ceux des $\mathcal{P}_i$. On remarque que les prix de chaque produit ont tendance à augmenter avec le temps car les clients sont prêts à payer plus cher quand on s'approche du départ de l'avion.

En plus d'une bonne borne que permet d'obtenir cet algorithme pour le Branch and Bound quand on fixera certains prix et que l'on résoudra le problème sans contraintes sur les prix restants, il nous permet aussi d'avoir une idée du revenu maximal en moyenne que l'on pourrait espérer obtenir pour un vol donné. Il permet ainsi d'évaluer l'efficacité d'une stratégie future en comparant le revenu moyen apporté par cette stratégie à ce revenu. 
Dans la Figure \ref{solution_continue}, vous trouverez la solution du problème relâché appliqué à chaque scénario dont on avait les données.

\begin{figure}[!h]
\centering
\begin{tabular}{ccc}
  \hline
  Scénario & Solution du problème continu & Nombre de passagers\\
  \hline
    1 & 76900.39 & 180 \\
    2 & 51678,19 & 180 \\
    3 & 19405,23 & 180 \\
    4 & 42077,20 & 180 \\
    5 & 58189,07 & 180 \\
  \hline
\end{tabular}
\caption{Revenu maximal pour chaque scénario que l'on peut obtenir sans contrainte sur les prix en résolvant le problème relâché avec un avion de capacité $C=180$.}
\label{solution_continue}
\end{figure}

La résolution de ce problème relâché aboutit pour chaque scénario à une saturation parfaite de l'avion. Ceci découle directement du fait de la relaxation de la contrainte sur les prix, on verra ainsi par la suite que cette constatation ne sera pas conservée lors de la résolution et de l'évaluation du problème initial.

\paragraph{Remarque}
Si l'on note $f_{i,t}(p) = p\,\alpha_{i,t}\,e^{-\beta_{i,t}\,p}$, cette fonction correspond au revenu individuel apporté par le produit $i$ au temps $t$ si on le fixe au prix $p$. $f_{i,t}$ est croissante jusqu'à son maximum en $\frac{1}{\beta_{i,t}}$ puis est décroissante. L'égalité $\mu^*\,f'(\mu^*) = 0$ décrit alors le fait que soit la contrainte de capacité est vérifiée (si $f'(\mu^*) = 0$) soit chaque prix correspond à la maximisation sans contrainte du revenu individuel $f_{i,t}$ (si $\mu^* = 0$ car $p_{i,t}^* = \mu^* + \frac{1}{\beta_{i,t}}$). \\

\subsubsection{Parcours de l'arbre et problèmes rencontrés}\label{idees_amelioration}
\label{}

Malgré la borne précédente qui permettait assez tôt de détecter les noeuds non satisfaisants pour la recherche dans l'arbre de Branch and Bound, l'algorithme ne convergeait pas après plusieurs heures de travail. Pour obtenir un temps raisonnable, nous avons identifié d'autres moyens permettant de couper de nombreuses branches de l'arbre et nous avons introduit une nouvelle hypothèse sur la solution recherchée.

\begin{itemize}
    \item On teste la pertinence de la branche que l'on parcours actuellement en regardant si la contrainte de capacité est vérifiée si tous les prix encore non fixés sont choisis égaux au prix maximal, induisant ainsi une demande minimale. Si ce n'est pas le cas, cela signifie qu'on ne vérifiera jamais la contrainte de capacité quelque soit les prix que l'on fixera par la suite. On détecte ainsi beaucoup plus haut dans l'arbre les problèmes de capacité futurs, ce qui nous permet d'éviter d'explorer une grande partie de cet arbre pour rien.
    \item Augmenter $p$ permet de libérer de la place pour les autres produits et les autres temps. Ainsi, si tous les autres prix sont fixés, on a tout intérêt à augmenter $p$ tant que cela permet d'augmenter le revenu individuel $f_{i,t}$. En effet, cela revient à augmenter le revenu global tout en maintenant la vérification de la contrainte de capacité car de la place est libérée.
    Le graphique en Figure \ref{fig:prix_non_optimaux} illustre comment on utilise cette constatation afin d'éliminer d'entrée des prix non optimaux à $t$ et $i$ fixés.
    \begin{figure}[!h]
        \begin{center}
        \subfigure{%
          \includegraphics[width=6.0cm]{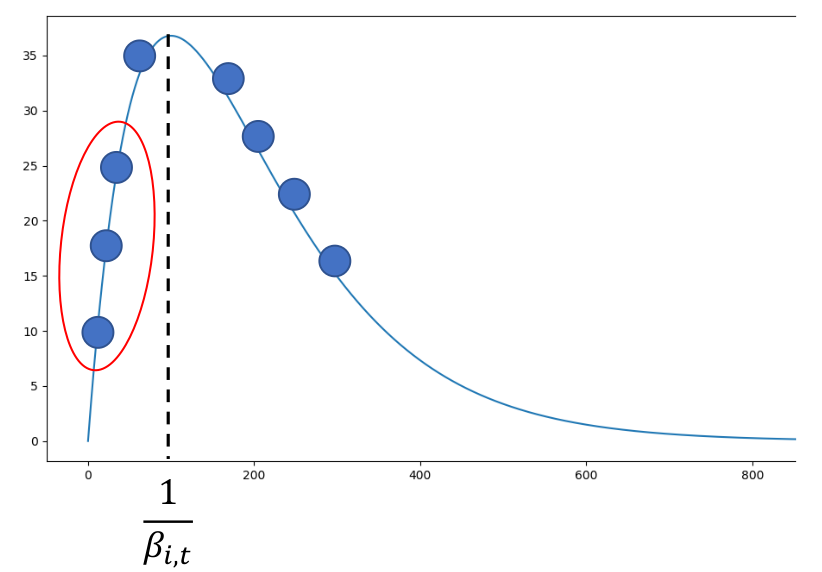}
          \label{fig:sub1}
        }
        \subfigure{%
          \includegraphics[width=6.0cm]{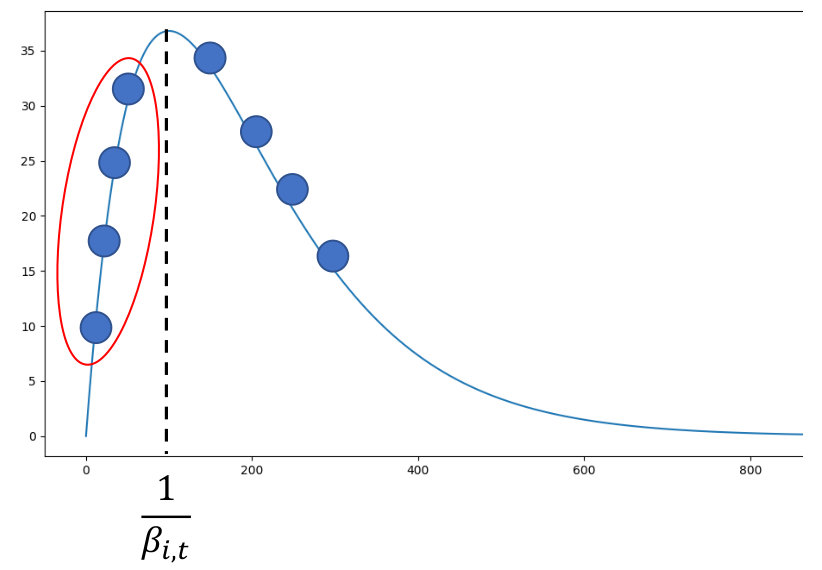}
          \label{fig:sub2}
        }
        \end{center}
        \caption{Pour le produit et le temps donné, les prix entourés sont directement éliminés car on peut à chaque fois trouver un prix plus élevé qui augmente le revenu. On remarque par ailleurs que le fait que les prix évoluent dans un ensemble discret fait que l'on sait que le dernier prix accepté sera un des deux points voisins de $\frac{1}{\beta_{i,t}}$ sans savoir a priori lequel des deux cela sera.}
        \label{fig:prix_non_optimaux}
    \end{figure}
    \item La dernière idée que nous avons utilisée ajoutait une hypothèse supplémentaire sur le résultat que l'on cherche. En supposant que l'on recherche des prix qui croissent quand on se rapproche du départ de l'avion, ce qui ne semble pas être trop restrictif au vu des résultats que nous avons eu précédemment, on peut de nouveau couper un grand nombre de branches dans l'arbre que l'on parcours.
\end{itemize}

Cependant, malgré l'ensemble de ces nouvelles améliorations qui permettaient de retirer un grand nombre de branches dans l'arbre du Branch and Bound, nous n'obtenions toujours pas de résultat après plusieurs heures de calcul. C'est pourquoi nous avons alors décidé de changer d'approche pour la résolution de ce problème.

\subsection{Une heuristique gloutonne}
\subsubsection{Algorithme de base}
Au lieu de parcourir l'ensemble de l'arbre du Branch and Bound, nous avons finalement opté pour un algorithme glouton, plus rapide à s'exécuter mais dont la qualité du résultat n'est pas garantie.

Le principe de l'algorithme est le suivant. Lorsque l'on se situe à un certain noeud de l'arbre du Branch and Bound, on s'intéresse à l'ensemble des fils du noeud courant. Chaque fils correspond à une certaine valeur de prix pour le même produit $i$ au même temps $t$. La résolution du relâché continu sur chaque fils de ce noeud nous donne une borne maximale atteignable avec des prix continus pour les prix non fixés. On poursuit alors la progression vers le fils où cette borne est maximale. Si au bout d'un moment ce parcours induit une violation de la contrainte de capacité, on remonte au noeud précédent et on parcours le fils donnant la deuxième meilleure borne maximale. Dans le cas contraire, on ne remontera jamais à un noeud précédent, chaque choix étant considéré comme définitif s'il n'induit pas de problème de capacité. Il s'agit donc d'un algorithme glouton qui s'arrêtera dès la première atteinte d'une feuille.

Des exemples de prix obtenus pour un avion de capacité $C = 180$ sont donnés à la Figure \ref{glouton_sc1} pour le scénario 1.

\begin{figure}[!h]
\centering
\begin{tabular}{cccc}
\toprule
{} $t$ &      Produit 1 &      Produit 2 &      Produit 3 \\
\midrule
0  &  700 &  350 &  250 \\
1  &  700 &  350 &  250 \\
2  &  700 &  350 &  250 \\
3  &  700 &  350 &  250 \\
4  &  700 &  350 &  250 \\
5  &  700 &  350 &  250 \\
6  &  700 &  350 &  250 \\
7  &  700 &  350 &  250 \\
8  &  633 &  350 &  250 \\
9  &  633 &  350 &  250 \\
10 &  567 &  350 &  250 \\
11 &  567 &  350 &  250 \\
12 &  567 &  350 &  250 \\
13 &  500 &  310 &  236 \\
14 &  500 &  310 &  236 \\
15 &  500 &  310 &  236 \\
16 &  500 &  310 &  236 \\
17 &  500 &  310 &  236 \\
18 &  500 &  310 &  236 \\
19 &  500 &  290 &  236 \\
20 &  500 &  290 &  236 \\
21 &  500 &  290 &  236 \\
22 &  433 &  290 &  236 \\
23 &  433 &  290 &  236 \\
24 &  433 &  290 &  236 \\
25 &  433 &  270 &  236 \\
26 &  433 &  270 &  236 \\
27 &  433 &  270 &  250 \\
28 &  433 &  290 &  250 \\
29 &  433 &  310 &  221 \\
\bottomrule
\end{tabular}
\caption{Prix fixé pour chaque produit et à chaque pas de temps du scénario 1 par l'algorithme glouton.}
\label{glouton_sc1}
\end{figure}

Le revenu alors obtenu est de $74787,38$. On remarque que les prix ont encore globalement tendance à augmenter à l'approche du départ de l'avion, mais la règle n'est pas exacte au vu des prix les plus éloignés du départ. Les revenus obtenus pour l'ensemble des scénarios avec comparaison à la borne maximale de la résolution continue sont donnés Figure \ref{glouton_scenes}.

\begin{figure}[!h]
\centering
\begin{tabular}{cccc}
\toprule
{} Scénario &      Revenu obtenu &      Borne, solution continue &      Passagers \\
\midrule
1  &  74787.38 &  76900.39 &  179.99 \\
2  &  51525.03 &  51678.19 &  179.71 \\
3  &  19332.18 &  19405.23 &  179.78 \\
4  &  41971.02 &  42077.20 &  179.68 \\
5  &  57869.83 &  58189.07 &  179.95 \\
\bottomrule
\end{tabular}
\caption{Résultats obtenus pour chaque scénario pour un avion de capacité $C= 180$.}
\label{glouton_scenes}
\end{figure}

On remarque qu'on ne perd pas trop par rapport à la solution optimale et que l'on arrive à chaque fois à saturer presque entièrement l'avion.

\paragraph{Remarques}
Pour cet algorithme, on utilise de plus les éléments de simplification de la Section \ref{idees_amelioration} qui n'induisent pas d'hypothèse sur les résultats, à savoir la suppression en amont des prix individuels non optimaux ainsi que la détection des problèmes de capacité futurs.

\subsubsection{Adaptation de la capacité pour des données réelles}
Nous avons ainsi obtenu une solution approchée a priori bonne du problème posé en Section \ref{modelisation_mathematique} en supposant que les données en moyenne étaient des données réelles. Nous revenons maintenant sur cette hypothèse afin de confronter notre algorithme à la réalité. A un pas de temps donné, on résout le problème sur l'ensemble des temps restants en supposant que l'on connaît la demande future comme on le faisait précédemment. Pour le pas de temps en cours, on fixe alors les premiers prix sortis par l'algorithme pour ce pas de temps. On regarde ensuite l'arrivée effective des passagers sur ce pas de temps. Chaque passager arrive en voulant acheter un certain produit à un certain prix. Si le prix indiqué lui convient, il achète le billet d'avion, sinon il ne l'achète pas. On a alors le revenu effectif et la demande effective associée à ce pas de temps. On adapte alors la capacité de l'avion restante à ces données réelles. Puis on refait la même chose sur le pas de temps suivant.

\paragraph{Remarque}
Les données réelles mentionnées précédemment sont en faîtes issues de simulations que nous avons implémentées à partir des mêmes données sur lesquelles nous travaillions depuis le début. On suppose en fait que le nombre de passagers à un instant donné pour un produit donné suit une loi de poisson de paramètre $Q_{i,t}$. Nous avons ensuite supposé que le prix maximum accepté $P_{i,t}^{max}$ pour chacun de ces passagers à ce temps $t$ et pour ce produit $i$ suit une loi de densité $g_{i,t}$ définie sur $[p_i^{min};+\infty[$ par :
$$
    g_{i,t}(p) = \frac{1}{s_{i,t}}\,\exp\left({-\frac{p-p_i^{min}}{s_{i,t}}}\right) \quad ; \quad s_{i,t} = \frac{p_i^{min}\,(F_{i,t}-1)}{\ln(2)}
$$

De telle sorte que la probabilité $p_{p_i^{min}\rightarrow p}$ qu'un individu parmi les $Q_{i,t}$ initiaux achète tout de même son billet d'avion si le prix est finalement $p$ soit :

$$
    p_{p_i^{min}\rightarrow p} = \mathbb{P}(p \leq P_{i,t}^{max}) = \int_{p}^{+\infty}\frac{1}{s_{i,t}}\,\exp\left({-\frac{q-p_i^{min}}{s_{i,t}}}\right) \, \mathrm dq = \exp\left(-\frac{\ln(2)}{F_{i,t}-1}\,\left(\frac{p}{p_i^{min}}-1\right)\right)
$$

Ce qui donne bien l'expression annoncée dans la section \ref{modelisation_mathematique} sur la modélisation mathématique de notre problème. La simulation de ces différentes lois nous permet alors d'obtenir la simulation d'arrivée réelle de passagers souhaitée.

Les résultats en moyenne obtenus pour chaque scène avec adaptation de la capacité de l'appareil en fonction de l'arrivée réelle des passagers sont présents dans la Figure \ref{resultats_moyens_reels}.

\begin{figure}[!h]
\centering 
\begin{tabular}{ccccc}
\toprule
{} Sc &  Revenu &      $\sigma_{revenu}$  &  Demande &  $\sigma_{demande}$ \\
\midrule
1 &                        72 351 &                  3 745 &                176 &             6 \\
2 &                        47 904 &                  4 037 &                146 &            10 \\
3 &                        17 666 &                  1 930 &                121 &             8 \\
4 &                        35 966 &                  3 049 &                124 &             8 \\
5 &                        53 125 &                  4 949 &                141 &            10 \\
\bottomrule
\end{tabular}
\caption{Moyenne et écart-type \textbf{empiriques} (100 itérations) du revenu et de la demande en fonction du scénario pour l'algorithme glouton avec une capacité $C=180$.}
\label{resultats_moyens_reels}
\end{figure}

On voit que la confrontation à une arrivée réelle de passagers nous fait perdre en terme de revenu et de saturation de l'avion. Afin de voir si cette solution est tout de même intéressante, nous allons comparer cet algorithme à d'autres plus classiques dans la section \ref{approche_classique}.

\section{Approche plus classique: EMSRB sur une transformation MR}\label{approche_classique}
Nous présentons dans cette partie une deuxième approche basée sur l'article \cite{article}. Cet article suggère une transformation des prix et des demandes dans le cadre d'un marché non segmenté (un client achètera le produit le moins cher disponible si celui-ci est dans son budget) afin de rendre les produits indépendants. 

\subsection{Description}
\paragraph{Notations}
Dans toute cette partie on notera: $f_i$ le prix lié au produit $i$ avec $f_1 > f_2 >\ldots> f_n$, $d_i$ la demande prête à payer au maximum $f_i$, $Q_k = \sum \limits_{j=1}^k d_j$ la demande agrégée des $k$ tarifs les plus chers, soit le nombre de ventes si le produit $k$ est le moins cher disponible, $TR_k = \sum \limits_{j=1}^k d_j f_j$ le revenu total si le produit $k$ est le moins cher disponible et $MR_k = \frac{TR_k - TR_{k-1}}{Q_k - Q_{k-1}}$ le revenu marginal du produit $k$. Dans le cas de notre algorithme dans lequel il existe également un découpage temporel, on rajoutera un indice $t$ désigne le pas de temps auquel on se trouve.

\paragraph{}
Étudions l'effet de l'ouverture de la classe $k$ dans l'hypothèse où les passagers achètent le produit le moins cher disponible.
L'ouverture de la classe $k$ va générer une augmentation de la demande de $d_k$ mais ne va pas entraîner une hausse du revenu de $d_k f_k$. En effet, les passagers qui allaient acheter le produit $f_{k-1}$ vont alors acheter le produit $f_k$ ce qui va entraîner une perte de revenu de $Q_{k-1} (f_{k-1} - f_k)$ due au "buy-down". Le revenu incrémental suite à l'ouverture de la classe k est donc $d_k f_k - Q_{k-1} (f_{k-1} - f_k)$. On peut le voir différemment en disant qu'au lieu de recevoir le revenu individuel $f_k$ en ouvrant la classe $k$ on obtient un revenu individuel ajusté $f'_k = d_k f_k - Q_{k-1} (f_{k-1} - f_k) $. On peut vérifier qu'on a alors $f'_k = MR_k$. Le demande qui va recevoir ce revenu ajusté est $d'_k = Q_k - Q_{k-1}$.
Plus le nombre de sièges disponibles à bas coût est important, plus le revenu marginal décroît et peut même devenir négatif. Il n'est alors plus intéressant d'ouvrir la classe car elle fait perdre de l'argent au lieu d'en faire gagner du fait de l'effet buy-down. Cette transformation permet donc de contrer la non-segmentation entre les classes.

\paragraph{}
La puissance de cette transformation des tarifs et des demandes réside dans le théorème suivant.

\paragraph{Théorème}
La transformation marginale modifie la structure de coût et créé un modèle équivalent dans lequel les produits sont indépendants.

\paragraph{Remarque}
On peut vérifier que dans le cas de produits indépendants (segmentation) la transformation donne bien $f'_k = f_k$ et $d'_k = d_k$.

\paragraph{}
Détaillons maintenant le principe de l'algorithme.
\begin{enumerate}
\item Pour chaque famille de produits et à chaque pas de temps nous effectuons d'abord une transformation marginale sur l'agrégation des temps futurs similaire à celle mentionnée précédemment ce qui nous donne des produits indépendants. Les produits entre les différentes familles étant déjà indépendants, on obtient un structure totalement segmentée.
\item On trie ces nouveaux produits par tarif marginal $f'_{k,t}$ décroissant
\item On applique alors un EMSRb nous donnant la politique à appliquer à chaque temps $t$.
\item A chaque pas de temps, pour chaque famille, on met en vente le produit au plus faible tarif de la famille pour lequel il reste encore des sièges à attribuer. Lorsqu'une vente est effectuée, le nombre de place s'ajuste et si le nombre de siège pour un tarif tombe à 0, la classe est fermée et on propose le prix du dessus.
\end{enumerate}

\subsection{Résultats}

Voici les résultats moyennés pour 100 échantillons et une capacité de 180 personnes en fonction du scénario. Les résultats sont proches de ceux proposés par l'algorithme glouton.

\begin{figure}[!h]
\centering
\begin{tabular}{ccccc}
\toprule
{}Sc &  Rev. moyen &      $\sigma_{revenu}$  &  Demande &  $\sigma_{demande}$ \\
\midrule
1 &  69 746 &  5 659 &                165 &                  9 \\
2 &  45 830 &  4 580 &                135 &                 11 \\
3 &  18 551 &  2 263 &                157 &                 11 \\
4 &  38 169 &  3 193 &                168 &                  5 \\
5 &  50 557 &  4 825 &                161 &                  8 \\
\bottomrule
\end{tabular}
\caption{Moyenne et écart-type du revenu et de la demande en fonction du scénario pour l'algorithme MRT-EMSRb pour une capacité $c=180$}
\label{mrt}
\end{figure}

Afin de pouvoir juger de l'importance de la transformation MRT, nous avons également implémenté un EMSRb classique en faisant comme si les tarifs étaient indépendants. Comme nous pouvons le voir sur la figure \ref{mrtvsclassique}, les revenus de l'EMSRb classique sont bien moins bon. La transformation MR permet donc bien de sécuriser davantage de sièges à haut revenu en fermant les classes non rentables. Cela s'explique par la non prise en compte de la dépendance entre les produits d'une même famille. De plus, on peut voir que la MRT permet de davantage remplir l'avion. L'hypothèse classique stipulant que les passagers arrivent par WTP ("willingness to pay" ou prix maximum accepté par un client) croissante est responsable de ce phénomène.  Du point de vue de l'algorithme classique il est toujours mieux de vendre le produit le plus cher, alors que  vendre un produit moins cher d'une autre famille peut se révéler en réalité plus rentable. Sur la figure \ref{cls} on peut voir que dans le cas classique, le produit de tarif $1000$ est plus attractif que le produit de tarif $800$, les sièges sont donc à réserver en priorité pour ce tarif. En revanche, si on prend en compte les différentes familles, on s'aperçoit que le tarif à $800$ est en réalité plus avantageux car il n'empêche pas les clients de consommer les places au tarif $1200$, leurs familles étant différentes.

\begin{figure}[!h]
\centering
\begin{tabular}{ccccccccc}
\toprule
{}  Famille &  Tarif &  Demande &   Q &      TR &     MR & Rés. Classique & Rés. MRT\\
\midrule
        1 &   1200 &       31 &  31 &  37 200 &  1 200 & 31 & 31 \\
        1 &   1000 &       11 &  42 &  42 000 &  454.5 & 11 & 0\\
        2 &    800 &       15 &  15 &  12 000 &    800 & 0 & 15\\
\bottomrule
\end{tabular}
\caption{Mise en évidence de l'utilité de la MR}
\label{cls}
\end{figure}

Supposons maintenant une arrivée de 57 passagers répartie selon la figure \ref{cls}. Si l'hypothèse de croissance des WTP est vérifiée alors les 57 places sont correctement attribuées avec les deux algorithmes et le revenu vaut $1200 \times 31 + 1000 \times 11 + 800 \times 15 = 53 908 $.

Cependant, si on suppose que les passagers arrivent dans un ordre aléatoire (puisque l'on se place à un temps fixé il n'y a pas de raison qu'il en soit autrement), certains passagers étant prêt à payer $1200$ peuvent arriver avant et remplir les places à $1000$. Les passagers arrivant par la suite et ayant une WTP de $1000$ n'auront alors plus de place.
\\ Prenons un exemple: Supposons qu'une demande de 60 arrive dans cet ordre et que nous disposons d'une capacité de 40:
\begin{itemize}
    \item 10 personnes intéressés par la famille 1 ayant un WTP de 1200
    \item 10 personnes intéressés par la famille 1 ayant un WTP de 1000
    \item 10 personnes intéressés par la famille 1 ayant un WTP de 800
    \item 10 personnes intéressés par la famille 1 ayant un WTP de 1200
    \item 10 personnes intéressés par la famille 1 ayant un WTP de 1200
    \item 10 personnes intéressés par la famille 1 ayant un WTP de 800
\end{itemize}
On peut alors vérifier que l'EMSRb classique va rapporter 35 000 pour 31 ventes tandis que la version avec MRT va rapporter 43 200 pour 39 ventes. Cet exemple illustre donc à la fois le meilleur remplissage et le meilleur revenu généré par le MRT-EMSRb.

\begin{figure}[!h]
\centering
\begin{tabular}{ccccc}
\toprule
{}Sc &  Rev. moyen MRT   &  Demande MRT &   Rev. moyen Classique &  Demande Classique \\
\midrule
1  &            69 746 &       165 &                  47 142 &                123 \\
2  &            45 830 &       135 &                  27 592 &                 73 \\
3  &            18 551 &       157 &                  10 135 &                 70 \\
4  &            38 169 &       168 &                  26 085 &                 93 \\
5  &            50 557 &       161 &                  35 038 &                104 \\
\bottomrule
\end{tabular}
\caption{Comparaison entre le MRT-EMSRb et un EMSRb classique pour une capacité $c=180$. La différence de revenu illustre l'importance de la prise en compte de la non-segmentation de la demande dans les modèles}
\label{mrtvsclassique}
\end{figure}

\section{Comparaison et évaluation des algorithmes}
Dans cette dernière partie, nous comparons les deux algorithmes que nous avons implémentés.
\begin{figure}[!h]
\centering 
\begin{tabular}{ccccc}
\toprule
{} Sc &  Rev. moyen Glouton   &  Demande Glouton &   Rev. moyen MRT &   Demande MRT \\
\midrule
1 &                        72 351 &                  176 &              69 746 &           165 \\
2 &                        47 904 &                  146 &              45 830 &           135 \\
3 &                        17 666 &                  121 &              18 551 &           157 \\
4 &                        35 966 &                  124 &              38 169 &           168 \\
5 &                        53 125 &                  141 &              50 557 &           161 \\
\bottomrule
\end{tabular}
\caption{Comparaison entre l'algorithme glouton et le MRT-EMSRb pour une capacité $c=180$.}
\label{gloutonvsmrt}
\end{figure}
\begin{itemize}
\item L'algorithme Glouton génère plus de revenus que le MRT pour les scénarios 1,3 et 5 qui sont caractérisés par une demande importante mais est moins performant lorsque la demande est faible (scénario 3 et 4).
\item L'algorithme Glouton a tendance à moins remplir l'avion mais génère en moyenne des revenus plus important sur ses ventes. Il a particulièrement du mal lorsque la demande est faible car il propose des tarifs élevés.
\item L'algorithme MRT tourne plus vite car la plupart du traitement s'effectue en amont. Aucun calcul ne se fait pendant l'arrivée de passager.
\item On a possibilité d'agir pas de temps par pas de temps en modifiant moyenne et variance de l'EMSRB en fonction du temps et du passé.
\item Le MRT permet de  prendre en compte la stochasticité de la demande.
\item L'algorithme glouton actuel ne permet pas de modifier les tarifs entre deux pas de temps, il y a donc encore de la marge pour augmenter les bénéfices avec cette méthode.
\end{itemize}

Cependant, les deux algorithmes ne sont pas très robustes à une estimation erronée de la demande. Ils ne tiennent compte de l'information passée que par la capacité qui s'actualise mais ne changent pas la politique future. On peut par exemple imaginer la possibilité de modifier les moyennes et variances de la gaussienne modélisant la demande dans l'EMSRb dynamiquement en tenant compte de la demande passée. (Figure \ref{oups})

\begin{figure}[!h]
\centering 
\begin{tabular}{cccccc}
\toprule
{} Sc estimé & Sc réel   &  Rev. potentiel MRT &   Rev. MRT &   Rev. potentiel glouton &  Rev. glouton\\
\midrule
1 &  3    &                  18 551 &                  8 268 &              17 666 &           9 019 \\
3 &  1      &                69 746 &                  36 515 &              72 351 &           27 887 \\
\bottomrule
\end{tabular}
\caption{Estimation de la robustesse des algorithmes sur des estimations erronées. Le scénario 1 prévoit une demande bien supérieure à celle du scénario 3.}
\label{oups}
\end{figure}

Dans le cas d'un scénario 3 non anticipé le nombre de sièges attribués est en moyenne de 19 et dans le cas d'un scénario 1 mal estimé l'avion est complet à chaque fois. On remarque alors que les performances des algorithmes sur des scénarios réels s'inversent dans le cas d'une mauvaise estimation. Ainsi, dans la pratique, si on estime une forte demande, il faudra privilégier l'algorithme glouton peu importe la précision de l'estimation et au contraire le MRT est à privilégier lorsqu'on s'attend à une faible demande.

\section{Conclusion}
Dans ce projet, nous avons élaboré des algorithmes de Revenue Management cherchant à s'affranchir de quelques hypothèses classiques notamment en prenant en compte la non-segmentation entre les différentes classes et l'élasticité-prix de la demande. Nous avons mis en place deux algorithmes se basant sur des approches très distinctes qui nous donnent des résultats complémentaires. Ces modèles ne sont pas encore très robustes et pourraient être complexifiés pour mieux s'adapter à de mauvaise estimations de la demande. L'objectif serait ensuite de tenter de généraliser ces idées au cadre plus complexe du réseau.
\bibliographystyle{abbrv}
\bibliography{main}

\end{document}